\documentclass{article}
\PassOptionsToPackage{numbers, compress}{natbib}   % before loading neurips_2026
\usepackage[preprint]{neurips_2026}
\usepackage[utf8]{inputenc}                         % allow utf-8 input
\usepackage[T1]{fontenc}                            % use 8-bit T1 fonts
\usepackage{microtype}                              % microtypography
\usepackage{natbib}
\usepackage{color}
\usepackage{xcolor,colortbl}                        % textcolor
\usepackage{longtable}
\usepackage{booktabs}                               % professional-quality tables
\usepackage{makecell}
\usepackage[export]{adjustbox}
\usepackage{multirow}
\usepackage{paralist}                               % compactitem
\usepackage{comment}                                % use comment
\usepackage{soul}                                   % use highlight command \hl{}
\soulregister\cite7
\soulregister\ref7
\soulregister\pageref7
\usepackage{enumerate}
\usepackage{url}
\usepackage{nth}                                    % nth command
\usepackage{courier}
\usepackage{threeparttable}
\usepackage{footnote}
\usepackage{listings}
\usepackage[inline]{enumitem}
\usepackage{verbatim}
\usepackage{etoolbox}                               % commands \newtoggle, \toggletrue, \iftoggle
\usepackage{pifont}                                 % ding commend
\usepackage[figuresright]{rotating}
\usepackage{moresize}
\usepackage{fontawesome}

\usepackage{amsmath}
\usepackage{amssymb}
\usepackage{amsfonts}                               % blackboard math symbols
\usepackage{amsthm}
\usepackage{bbding}                                 % checkmark
\usepackage{physics}                                % dv, pdv
\usepackage[mathcal]{eucal}
\usepackage{mathrsfs}
\usepackage{bm}                                     % bm command
\usepackage{blkarray}                               % to support matrix
\usepackage{nicefrac}                               % compact symbols for 1/2, etc.

\usepackage{wrapfig}
\usepackage{graphicx}                               % include pdf figures
\usepackage{eqparbox}                               % for equal height boxes
\usepackage{subcaption}
\usepackage{caption}
\usepackage{cleveref}
\usepackage{tikz}                                          
\usepackage{circuitikz}
\usetikzlibrary{patterns,snakes}
\usetikzlibrary{positioning,calc,fit,decorations.pathmorphing,shapes.geometric, shapes.gates.logic.US, calc}
\usetikzlibrary{arrows,arrows.meta,decorations.markings,shapes,shapes.arrows}
\usetikzlibrary{decorations,decorations.pathreplacing}
\usetikzlibrary{backgrounds}
\usepackage{filecontents}                           % support to pgfplots
\usepackage{pgfplots}
\usepackage{pgfplotstable}
\usepgfplotslibrary{groupplots}
\usepackage{scalefnt}
\pgfplotsset{compat=newest}

\usepackage{algorithm}
\iffalse
\usepackage{algpseudocode}                          % new algorithm package
\algrenewcommand\textproc{\texttt}
\makeatletter
\let\OldStatex\Statex
\renewcommand{\Statex}[1][3]{%
  \setlength\@tempdima{\algorithmicindent}%
  \OldStatex\hskip\dimexpr#1\@tempdima\relax
}
\makeatother
\else
\usepackage{algorithmic}
\fi

\crefname{mytheorem}{Theorem}{Theorems}
\crefname{mylemma}{Lemma}{Lemmas}
\crefname{myclaim}{Claim}{Claims}
\crefname{myproperty}{Property}{Properties}
\crefname{mycorollary}{Corollary}{Corollaries}

\newcommand{\cbit}{\begin{compactitem}}
	\newcommand{\ceit}{\end{compactitem}}
\newcommand{\cben}{\begin{compactenum}}
	\newcommand{\ceen}{\end{compactenum}}

\newcommand{\minisection}[1]{\vspace{.01in}\noindent{\textbf{#1}}.}

\usepackage{tcolorbox}
\tcbuselibrary{skins,breakable}
    {\endtcolorbox}
    {\endtcolorbox}

\definecolor{CUHKorange}{RGB}{244,106,18} %F47012
\definecolor{CUHKblue}{RGB}{0,111,190}    %006FBE
\definecolor{CUHKgreen}{RGB}{0,127,128}   %007F80
\definecolor{CUHKred}{RGB}{228,46,36}     %E42E24
\definecolor{CUHKyellow}{RGB}{198,148,34} %C69422
\definecolor{CUHKdark}{RGB}{114,44,114}   %722C72
\definecolor{CUHKmiddle}{RGB}{144,44,144} %902C90
\definecolor{CUHKlight}{RGB}{167,44,167} 
\definecolor{CUHKpurple}{RGB}{117,15,109}
\definecolor{CUHKgold}{RGB}{221,163,0}
\definecolor{CUHKribbon}{RGB}{244,223,176}
\definecolor{CUHKblack}{RGB}{34,24,21}

\graphicspath{{./figures/}{../}}

\title{\textsc{FocuSFT}: Bilevel Optimization for Dilution-Aware Long-Context Fine-Tuning}

\author{
    Zehua Pei$^1$,  
    Hui-Ling Zhen$^2$, 
    Xianzhi Yu$^2$, 
    Sinno Jialin Pan$^1$,
    Mingxuan Yuan$^2$, 
    Bei Yu$^1$\\
    $^1$The Chinese University of Hong Kong \quad
    $^2$Huawei Technologies Co., Ltd
}

\begin{document}
\maketitle

\begin{abstract}
Large language models can now process increasingly long inputs, yet their ability to effectively use information spread across long contexts remains limited.
We trace this gap to how attention budget is spent during supervised fine-tuning (SFT) on long sequences: positional biases and attention sinks cause the model to allocate most of its attention to positionally privileged tokens rather than semantically relevant content.
This \emph{training-time attention dilution} (the starvation of content tokens in the attention distribution) weakens the gradient signal, limiting the model's ability to learn robust long-context capabilities.
We introduce \textsc{FocuSFT}, a bilevel optimization framework that addresses this problem at training time.
An inner loop adapts lightweight fast-weight parameters on the training context to form a parametric memory that concentrates attention on relevant content, and the outer loop performs SFT conditioned on this sharpened representation.
Both loops apply bidirectional attention over context tokens while preserving causal masking for responses, reducing the causal asymmetry that gives rise to attention sinks and aligning inner-outer behavior.
On BABILong, \textsc{FocuSFT} improves accuracy by up to +14pp across 4K--32K context lengths; on RULER, it raises CWE aggregation from 72.9\% to 81.1\% at 16K; and on GPQA with agentic tool use, it yields a 24\% relative gain in pass@1.
Attention analysis shows that \textsc{FocuSFT} reduces attention sink mass by 529$\times$ and triples context engagement during training.
Code: \url{https://github.com/JarvisPei/FocuSFT}
\end{abstract}

\section{Introduction}
\label{sec:intro}

Many applications of large language models (LLMs) rely on long-context capabilities: analyzing scientific corpora, synthesizing documents, maintaining coherent multi-turn dialogues, and reasoning over large code repositories~\cite{jimenez2023swe, bai2024longbench,pei2025scope}.
Recent advances in positional encoding, distributed training, and architectural design have expanded context windows by orders of magnitude~\cite{dubey2024llama, liu2023scaling, ding2024longrope, team2024gemini}.
On the surface, the long-context problem appears largely solved: modern frontier models can ingest far more tokens than ever before.

However, a growing body of empirical evidence exposes a fundamental gap between context capacity and context utilization.
RULER~\cite{hsieh2024ruler} showed that many models scoring highly on simple needle-in-a-haystack retrieval suffer large performance drops as task complexity increases, with most models failing to maintain effective performance at their advertised context lengths.
The ``Lost in the Middle'' phenomenon~\cite{liu2024lost} revealed a U-shaped accuracy curve: LLMs attend well to the beginning and end of the input but systematically neglect content in the middle.
These findings point to a common conclusion: a larger context window does not imply a larger reliable working memory.
For instance, Qwen2.5-7B~\cite{yang2024qwen2} supports 128K tokens, yet already struggles on reasoning tasks at 4K--32K, well within its native capacity.
The problem is particularly acute in agentic settings, where the context comprises complex multi-turn dialogues including system prompts, user instructions, tool calls and their outputs, and prior assistant responses.
In such scenarios, the model must attend to relevant information dispersed across structurally heterogeneous turns, making it especially vulnerable to positional biases.

The root causes are well studied at inference time: positional biases direct attention toward the beginning and end of the context~\cite{liu2024lost, hsieh2024found}, and attention sinks consume a large share of the budget on a handful of initial tokens~\cite{xiao2023efficient}.
We refer to the resulting starvation of content tokens as \emph{attention dilution} (formalized in \Cref{sec:attn_review}).
Existing remedies overwhelmingly target inference (positional calibration~\cite{hsieh2024found}, dynamic scaling~\cite{ye2026dysco}, test-time training~\cite{bansal2025let, tandon2025end}) or require pretraining from scratch~\cite{child2019generating, ye2024differential}.

What remains largely unexplored is whether the fine-tuning procedure itself contributes to this gap.
We present evidence that it does: during standard SFT on a long sequence, the same biases and sink patterns govern the forward pass that produces the training loss.
The gradient signal is computed from representations where most attention goes to positionally privileged tokens rather than content.
Longer training sequences may reinforce rather than correct these patterns, creating a vicious cycle in which training-time dilution leads to poor long-context learning.

\begin{figure}[t]
  \centering
  \includegraphics[width=0.9\textwidth]{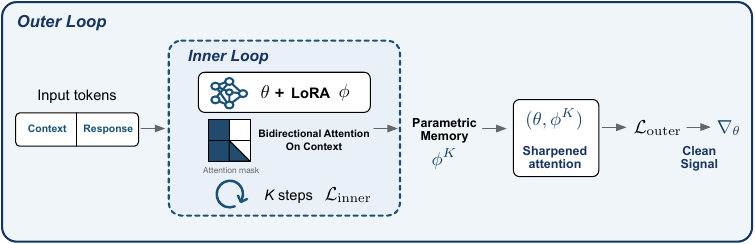}
  \caption{Overview of \textsc{FocuSFT}.
  Each training step is a bilevel optimization: the outer loop (full box) performs standard SFT on response tokens, while an inner loop (dashed box, nested inside) first adapts lightweight LoRA fast weights $\bm{\phi}$ on the context with bidirectional attention, forming a parametric memory $\bm{\phi}^{(K)}$ that concentrates attention on relevant content.
  The outer loop then computes the SFT loss conditioned on this sharpened representation, yielding a more informative gradient signal for updating $\bm{\theta}$.
  The fast weights are discarded after each step; only $\bm{\theta}$ persists.}
  \label{fig:overview}
\end{figure}

We propose \textsc{FocuSFT}, a dilution-aware fine-tuning framework that breaks this cycle through bilevel optimization (\Cref{fig:overview}).
In the inner loop, lightweight fast-weight adapters~\cite{hinton1987using, ba2016using} perform a small number of gradient steps on the training context, forming a parametric memory that concentrates attention on relevant content.
The outer loop then performs standard SFT conditioned on the sharpened representations produced by this memory, so that the gradient signal reflects actual context content rather than a diluted approximation.
To mitigate the sink mechanism, both loops apply bidirectional attention over context tokens while preserving causal masking for responses~\cite{du2022glm}: when all context tokens can attend to each other, the asymmetric visibility that drives sinks is reduced.
A key design principle, inner-outer consistency, aligns both loops to share the same attention structure and objective, so that the sharpened representations remain compatible with how the model processes context during training and inference.

Our main contributions are as follows:
\begin{itemize}[leftmargin=1.5em, itemsep=2pt, topsep=2pt]
  \item We identify training-time attention dilution (the starvation of content tokens due to positional biases and learned sinks) as a previously under-explored bottleneck for long-context learning, and characterize the vicious cycle between diluted training signals and poor long-context utilization.
  \item We propose \textsc{FocuSFT}, a bilevel optimization framework in which an inner loop constructs parametric memory via fast-weight adaptation and an outer loop performs SFT conditioned on the sharpened representation. Both loops employ bidirectional context attention that reduces the causal asymmetry linked to attention sinks, unified by an inner-outer consistency principle.
  \item We demonstrate consistent improvements across benchmarks: up to +14pp on BABILong at 4K--32K, +8.2pp on RULER CWE aggregation at 16K, and +3.8pp pass@1 on GPQA agentic reasoning. Attention analysis shows a 529$\times$ reduction in sink mass and 3.1$\times$ higher context engagement.
\end{itemize}

\section{Preliminaries and Motivation}
\label{sec:motivation}

\subsection{Attention Mechanisms and Long-Context Failure Modes}
\label{sec:attn_review}

We briefly review scaled dot-product attention and the failure modes that arise as context length grows.

Given a sequence of $T$ tokens with hidden representations $\{\mathbf{h}_i\}_{i=1}^T \in \mathbb{R}^d$, each transformer layer computes query, key, and value projections $\mathbf{q}_i = W_Q \mathbf{h}_i$, $\mathbf{k}_j = W_K \mathbf{h}_j$, $\mathbf{v}_j = W_V \mathbf{h}_j$, where $W_Q, W_K \in \mathbb{R}^{d_k \times d}$ and $W_V \in \mathbb{R}^{d_v \times d}$.
The attention logits $z_{i,j} = \mathbf{q}_i^\top \mathbf{k}_j / \sqrt{d_k}$ are normalized via softmax to yield attention weights:
\begin{equation}
  \alpha_{i,j} = \frac{\exp(z_{i,j})}{\sum_{\ell=1}^{T} \exp(z_{i,\ell})}, \qquad
  \mathbf{o}_i = \sum_{j=1}^{T} \alpha_{i,j}\, \mathbf{v}_j.
  \label{eq:attention}
\end{equation}
In autoregressive models, causal masking restricts the sum to $j \le i$.

\minisection{Positional attention bias}
LLMs exhibit a well-documented U-shaped positional bias~\cite{liu2024lost}: tokens at the beginning and end of the context receive systematically higher attention, irrespective of their relevance.
This effect is shaped by positional encoding schemes such as RoPE distance decay and reinforced by the training data distribution.
As a result, information placed in the middle of the context may be effectively invisible to the model, even when the context window comfortably accommodates it.

\minisection{Attention sinks}
Under causal masking, the first few tokens are the only positions visible to all subsequent tokens in the sequence.
Models learn to exploit this unique global visibility by using these initial tokens as attention sinks~\cite{xiao2023efficient}: destinations that absorb excess attention mass when no other token is strongly relevant.
This is not a defect but a functional mechanism: the model needs somewhere to place the probability mass that softmax forces it to allocate, and the globally visible initial tokens are a natural choice.
However, the consequence is that a substantial fraction of the attention budget is consumed by a handful of tokens that carry no semantic relevance.

\minisection{Attention dilution}
Together, positional bias and learned sinks cause the attention budget available for semantically relevant content to be \emph{diluted}: the model allocates most of its attention to positionally privileged tokens, leaving content tokens underattended.
We refer to this overall phenomenon as \emph{attention dilution}.

\subsection{Training-Time Attention Dilution: The Overlooked Bottleneck}
\label{sec:training_dilution}

\begin{figure}[t]
  \centering
  \begin{minipage}[t]{0.48\textwidth}
    \centering
    \includegraphics[width=\textwidth]{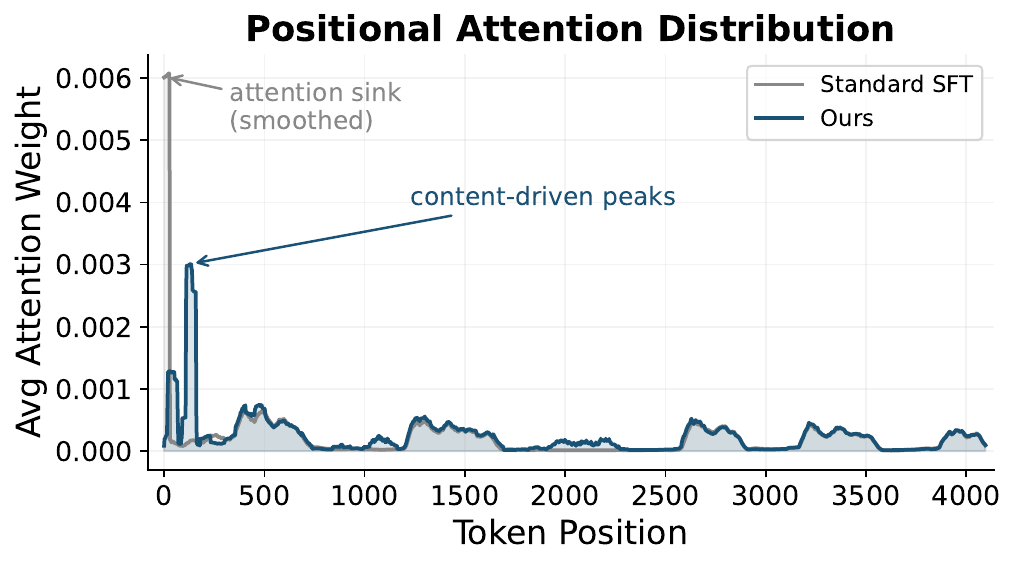}
    \subcaption{Positional attention distribution}
    \label{fig:positional_attn}
  \end{minipage}
  \hfill
  \begin{minipage}[t]{0.48\textwidth}
    \centering
    \includegraphics[width=\textwidth]{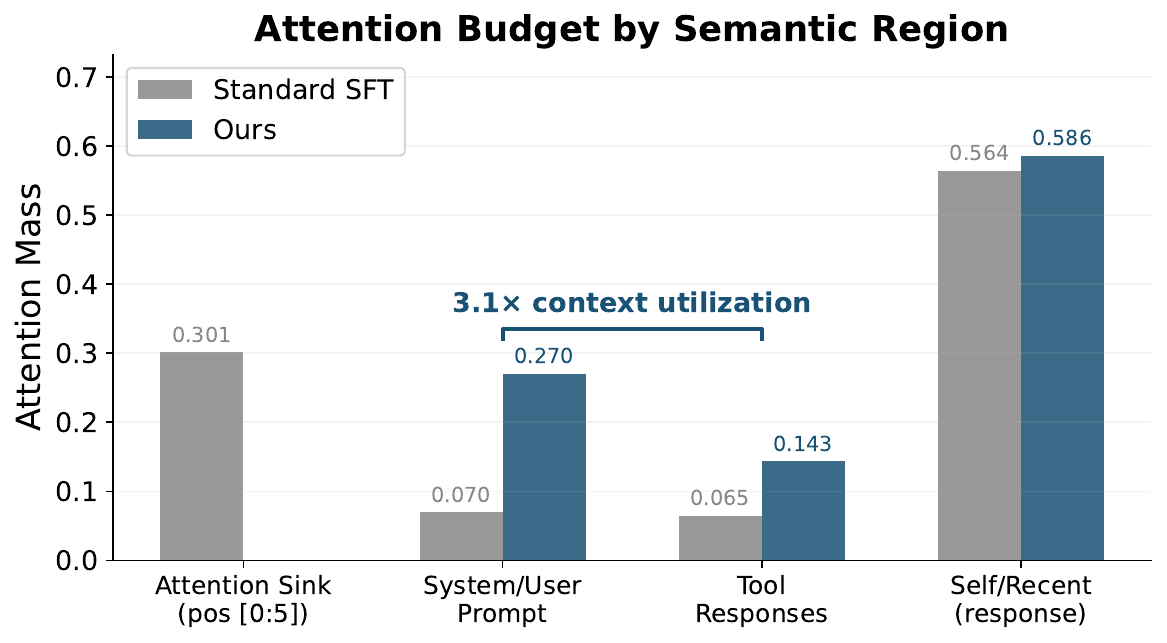}
    \subcaption{Attention budget by semantic region}
    \label{fig:budget_detail}
  \end{minipage}
  \caption{Training-time attention patterns on a 4096-token multi-turn agentic sample, comparing standard SFT and \textsc{FocuSFT} (response-query attention averaged across all 28 layers).
  (a) Standard SFT is dominated by the attention sink with near-zero attention elsewhere; \textsc{FocuSFT} produces content-driven peaks at meaningful positions.
  (b) Standard SFT directs only 13.5\% of attention to context content; \textsc{FocuSFT} achieves 41.4\% (3.1$\times$).
  % See also \Cref{fig:attn_heatmap} for raw attention matrices and \Cref{fig:sink_per_layer} for per-layer analysis.
  }
  \label{fig:training_dilution}
\end{figure}

\begin{figure}[t]
  \centering
  \includegraphics[width=\textwidth]{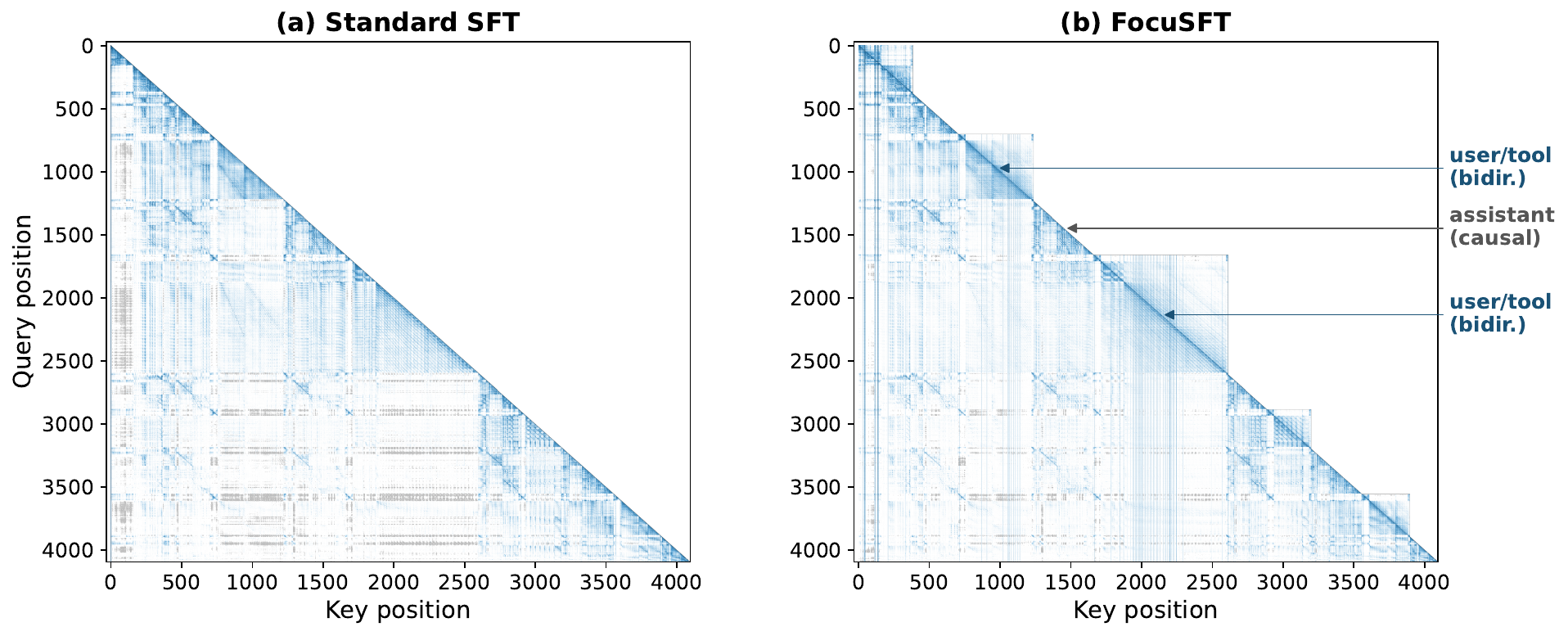}
  \caption{Attention heatmaps at a representative middle layer on a 4096-token multi-turn sample.
  (a)~Standard SFT: a bright column at initial tokens reveals the attention sink absorbing budget across all query positions.
  (b)~\textsc{FocuSFT}: the sink vanishes, revealing the multi-turn dialogue structure with bidirectional blocks for context turns and causal blocks for assistant responses.}
  \label{fig:attn_heatmap}
\end{figure}

Prior work has treated attention dilution primarily as an inference-time phenomenon.
However, inference-time attention is a product of the learned parameters, which are shaped by training-time attention.
We argue that a critical bottleneck lies in the training process itself, and provide empirical evidence in \Cref{fig:training_dilution,fig:attn_heatmap}.

Consider a standard SFT step on a long training sequence of length $T$.
During the forward pass, the attention mechanism computes weights $\alpha_{i,j}$ over the entire sequence.
Because of the patterns described above (positional bias and learned sinks), the output representation $\mathbf{o}_i$ at the prediction position is dominated by positionally privileged tokens rather than semantically relevant context.
The cross-entropy loss computed from this representation therefore provides a gradient signal that reflects a diluted view of the training data: the model sees the relevant information in its context window but cannot attend to it.

\Cref{fig:positional_attn} makes this concrete: under standard SFT, nearly all attention mass concentrates at position 0 with negligible weight elsewhere.
\Cref{fig:budget_detail} quantifies the consequence: the attention sink absorbs 30.1\% of the budget on just 5 tokens, while the entire context content (system/user prompt and tool responses combined) receives only 13.5\%.
\Cref{fig:attn_heatmap} further reveals the structural pattern: a bright sink column at the initial tokens dominates across all query positions, obscuring the underlying dialogue structure.
Over many training steps, the model converges to these attention patterns, failing to learn the sharp, content-specific focus required for reliable long-context utilization.
(For comparison, these figures also show the corresponding patterns under \textsc{FocuSFT}; we analyze these in \Cref{sec:attn_analysis}.)

This creates a vicious cycle:
\begin{center}
\begin{tikzpicture}[
  node distance=1.2cm,
  box/.style={draw, rounded corners=3pt, text width=2.6cm, minimum height=0.9cm,
              align=center, font=\footnotesize, inner sep=4pt, fill=white, line width=0.5pt},
  arr/.style={-{Stealth[length=4pt]}, line width=0.7pt, color=black!70},
  darr/.style={-{Stealth[length=4pt]}, line width=0.7pt, color=black!45, dashed},
]
  \node (A) [box, fill=black!5] {Attention dilution\\during training};
  \node (B) [box, right=of A]   {Shallow attention\\patterns learned};
  \node (C) [box, right=of B, fill=black!5] {Poor long-context\\utilization};
  \draw[arr] (A) -- (B);
  \draw[arr] (B) -- (C);
  \draw[darr] (C.south) -- ++(0, -0.55) -| node[below, midway, font=\scriptsize\itshape, text=black!55] {reinforces} (A.south);
\end{tikzpicture}
\end{center}

A natural response is to simply train on longer sequences.
However, longer sequences exacerbate rather than alleviate dilution: the attention sink absorbs an even larger share of the budget, and the model has more distractor tokens competing for what remains.
Empirically, models trained with longer context windows often show improved performance at short contexts but diminishing gains at the lengths they were trained on~\cite{hsieh2024ruler, kuratov2024babilong}.

These observations motivate a training-time approach that addresses two complementary aspects of the problem.
First, the root cause: under causal masking, the asymmetric visibility structure creates attention sinks that waste attention budget.
Bidirectional context attention addresses this asymmetry: when all context tokens can attend to each other, initial tokens are no longer uniquely privileged, reducing the pressure that drives the sink mechanism.
Second, addressing the mask structure alone is insufficient: the model must still learn to concentrate attention on semantically relevant content, which requires active guidance during training.
Bilevel optimization with an inner-loop parametric memory provides this guidance by sharpening the attention distribution, so that the outer-loop gradient signal better reflects actual context content.
In the next section, we present how \textsc{FocuSFT} realizes these ideas.

\section{Methodology}
\label{sec:method}

\subsection{Bilevel Optimization Framework}
\label{sec:method_overview}

Let $\bm{\theta}$ denote the base model parameters and $\mathbf{x} = [\mathbf{x}_{\text{ctx}};\, \mathbf{x}_{\text{resp}}]$ a training sequence of $T$ tokens, where $\mathcal{C}$ and $\mathcal{R}$ denote the sets of context and response token positions ($\mathcal{C} \cup \mathcal{R} = \{1, \ldots, T\}$).
Let $\bm{\phi}$ be a set of lightweight fast-weight parameters (LoRA adapters~\cite{hu2021lora}) that are re-initialized at each training step.
\textsc{FocuSFT} decomposes each step into two nested optimization levels:
\begin{equation}
  \min_{\bm{\theta}} \;\; \mathcal{L}_{\text{outer}}\!\bigl(\bm{\theta}, \bm{\phi}^{(K)}(\bm{\theta})\bigr),
  \qquad \text{where} \quad
  \bm{\phi}^{(K)} = \textsc{InnerLoop}\bigl(\bm{\phi}^{(0)},\, \bm{\theta},\, \mathbf{x}\bigr).
  \label{eq:bilevel}
\end{equation}
The inner loop runs $K$ gradient steps to adapt $\bm{\phi}$, producing $\bm{\phi}^{(K)}$; the outer loop then performs standard SFT on the response tokens conditioned on these adapted fast weights.
This structure is inspired by meta-learning~\cite{finn2017model, nichol2018first} and fast-weight mechanisms~\cite{hinton1987using, ba2016using}, repurposed to counteract attention dilution during training.

\subsection{Inner Loop: Parametric Memory}
\label{sec:inner_loop}

The fast weights $\bm{\phi}$ are LoRA adapters applied to the feed-forward network (FFN) of a selected subset of transformer layers, re-initialized to zero at each training step.
The inner loop performs $K$ gradient steps to minimize an adaptation loss $\mathcal{L}_{\text{inner}}$ on the response tokens, using the same next-token prediction objective as the outer loop:
$\mathcal{L}_{\text{inner}}(\bm{\theta}, \bm{\phi}; \mathbf{x}) = -\sum_{i \in \mathcal{R}} \log p_{\bm{\theta}, \bm{\phi}}(x_i \mid \mathbf{x}_{<i})$.
The update rule is:
\begin{equation}
  \bm{\phi}^{(k+1)} = \bm{\phi}^{(k)} - \eta_{\text{in}} \, \nabla_{\bm{\phi}} \mathcal{L}_{\text{inner}}\!\bigl(\bm{\theta}, \bm{\phi}^{(k)};\, \mathbf{x}\bigr),
  \qquad k = 0, \ldots, K{-}1,
  \label{eq:inner_update}
\end{equation}
where $\eta_{\text{in}}$ is the inner learning rate.
By optimizing the same response prediction objective, the inner loop forces $\bm{\phi}$ to encode context information that is directly useful for generating accurate responses.
After $K$ steps, the adapted $\bm{\phi}^{(K)}$ modifies the model's intermediate representations, indirectly reshaping the attention distribution in subsequent layers to better concentrate on the salient content of the current sample.

We adopt a first-order approximation~\cite{nichol2018first, rajeswaran2019meta}: the outer loss treats $\bm{\phi}^{(K)}$ as a constant, avoiding second-order derivatives through the inner-loop graph.
This reduces memory and compute overhead while preserving the core benefit.

\subsection{Outer Loop: SFT with Sharpened Attention}
\label{sec:outer_loop}

The outer loop performs a standard autoregressive forward pass using the combined parameters $(\bm{\theta}, \bm{\phi}^{(K)})$ and computes cross-entropy on the response token positions $\mathcal{R}$:
\begin{equation}
  \mathcal{L}_{\text{outer}}(\bm{\theta}, \bm{\phi}^{(K)}) = - \sum_{i \in \mathcal{R}} \log p_{\bm{\theta}, \bm{\phi}^{(K)}}(x_i \mid \mathbf{x}_{<i}).
  \label{eq:outer_loss}
\end{equation}
Only $\bm{\theta}$ receives gradient updates; $\bm{\phi}^{(K)}$ is discarded after each step.
Because the fast weights shift attention toward relevant context, the gradient signal reaching $\bm{\theta}$ better reflects actual content rather than a diluted approximation.

\subsection{Bidirectional Context Attention}
\label{sec:bidir_attn}

Following GLM-style attention~\cite{du2022glm}, we apply bidirectional attention over context tokens while preserving causal masking for responses.
The attention mask $M \in \{0, -\infty\}^{T \times T}$ is defined as:
\begin{equation}
  M_{i,j} =
  \begin{cases}
    0       & \text{if } i, j \in \mathcal{C}, \\
    0       & \text{if } i \in \mathcal{R} \text{ and } j \le i, \\
    -\infty & \text{otherwise.}
  \end{cases}
  \label{eq:bidir_mask}
\end{equation}
This mask is applied identically across all attention heads and in both the inner and outer loops.

Bidirectional context attention addresses the root cause of attention sinks identified in \Cref{sec:attn_review}: under causal masking, initial tokens are the only globally visible positions and absorb excess attention mass.
When all context tokens can attend to each other, this asymmetry vanishes and the sink mechanism becomes unnecessary.
This is particularly beneficial for the inner loop, where a complete view of the context enables more effective parametric memory formation.

\subsection{Inner-Outer Consistency}
\label{sec:consistency}

The effectiveness of \textsc{FocuSFT} depends on alignment between the two loops, a principle we call inner-outer consistency.
If the inner loop operates under conditions that differ from the outer loop (e.g., mismatched attention masks or objectives), the fast weights may encode representations that are incompatible with the outer loop, producing distortion rather than sharpening.

We enforce consistency along two dimensions.
\textbf{Objective}: the inner loop minimizes the same next-token prediction loss on response tokens as the outer loop (\Cref{eq:outer_loss}), so the fast weights are optimized to encode context representations that directly improve response generation.
\textbf{Attention pattern}: both loops use the same attention mask (\Cref{eq:bidir_mask}), with bidirectional attention over context tokens and causal masking for responses.
Because the fast weights are LoRA adapters on the same FFN layers used by the outer loop, the sharpened representations are directly compatible by construction.
The inner loop thus operates as a preview of the outer loop under identical conditions, forcing the fast weights to encode context representations that are directly compatible with the outer-loop objective.
At inference time, no inner-loop computation is required: the fine-tuned $\bm{\theta}$ is used with standard autoregressive decoding, and the bilevel training produces attention patterns that are more content-focused even under standard causal masking.

\section{Experiments}
\label{sec:exp}

We evaluate \textsc{FocuSFT} on long-context understanding benchmarks spanning synthetic reasoning, retrieval-aggregation, real-world QA, and agentic reasoning.

\subsection{Experimental Setup}
\label{sec:exp_setup}

\minisection{Base model and training data}
We use Qwen2.5-7B~\cite{yang2024qwen2} as the base model.
Training uses 3K multi-turn agentic SFT samples~\cite{yu2025demystify} with a maximum sequence length of 4096 tokens, trained for 5 epochs with an effective batch size of 32 (8 GPUs, gradient accumulation 4).
The outer-loop optimizer is AdamW~\cite{loshchilov2017decoupled} with learning rate $1 \times 10^{-5}$, cosine schedule with 10\% warmup, and weight decay 0.01.
All training uses BF16 mixed precision.

\minisection{Bilevel hyperparameters}
The inner loop performs $K{=}2$ gradient steps with learning rate 1.0 on LoRA~\cite{hu2021lora} adapters (rank 32, $\alpha{=}64$) applied to the FFN layers of the top 35\% of transformer layers.
Inner gradients are clipped at norm 1.0.
Full hyperparameter details are provided in \Cref{app:details}.

\minisection{Baselines}
We compare against: (1)~the pretrained Qwen2.5-7B without fine-tuning, and (2)~Standard SFT with identical data, model, and training budget but no bilevel optimization.
For ablations, we additionally test SFT with bidirectional context attention (no bilevel) and causal bilevel (bilevel without bidirectional context).

\begin{wrapfigure}{r}{0.46\textwidth}
  \vspace{-0.6cm}
  \centering
  \includegraphics[width=0.44\textwidth]{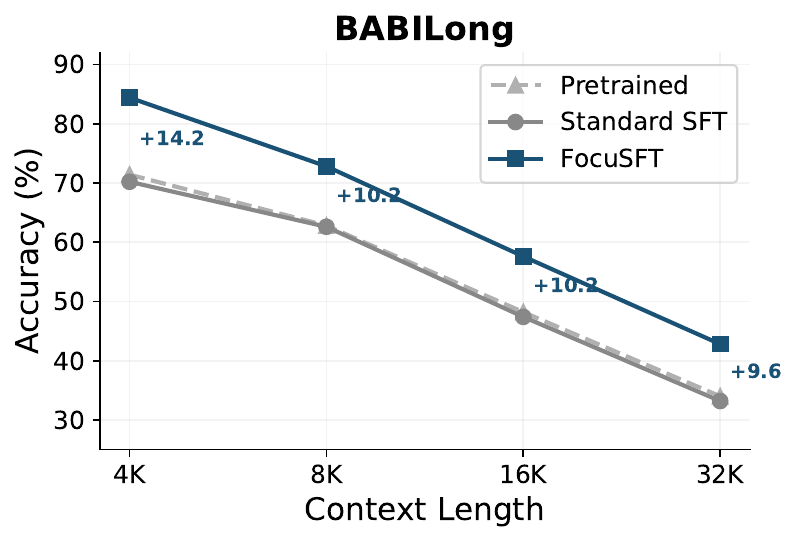}
  \caption{BABILong accuracy across context lengths.}
  \label{fig:babilong}
  \vspace{-0.2cm}
\end{wrapfigure}

\minisection{Evaluation benchmarks}
\textbf{BABILong}~\cite{kuratov2024babilong}: reasoning-in-a-haystack tasks at context lengths 4K--32K, testing fact retrieval and multi-hop reasoning within long narratives.
\textbf{RULER}~\cite{hsieh2024ruler}: multi-category benchmark covering retrieval (NIAH-MultiValue), aggregation (CWE), and multi-hop tracing (VT).
\textbf{LongBench}~\cite{bai2024longbench}: real-world QA tasks (HotpotQA, MultifieldQA, NarrativeQA, Qasper) at 8K context.
\textbf{GPQA}~\cite{rein2023gpqa}: graduate-level science reasoning (198 Diamond problems) evaluated with multi-turn agentic tool use via Open-AgentRL~\cite{yu2025demystify} ($n{=}32$ rollouts per problem).

\subsection{Main Results}
\label{sec:main_results}

\minisection{BABILong}
\Cref{fig:babilong} presents the main BABILong results.
\textsc{FocuSFT} outperforms Standard SFT by +14.2, +10.2, +10.2, and +9.6pp at 4K, 8K, 16K, and 32K respectively.
Standard SFT provides essentially no improvement over the pretrained model at any length, suggesting that naive fine-tuning with diluted attention fails to teach long-context reasoning.
\textsc{FocuSFT} maintains its advantage well beyond the 4K training length, demonstrating that dilution-aware training produces representations that generalize to longer sequences.
The gains are largest on multi-hop subtasks that require connecting dispersed facts across the context (\Cref{app:babilong_pertask}), the setting where attention dilution is most harmful.

\minisection{RULER}
\Cref{tab:ruler} shows per-task RULER results.
The improvement is most pronounced on CWE, where \textsc{FocuSFT} achieves 81.1\% vs.\ 72.9\% (+8.2pp) at 16K.
CWE requires aggregating information spread across the full context, consistent with the benefit of reduced attention dilution.
On NIAH-MV and VT, both methods perform near ceiling at shorter lengths; \textsc{FocuSFT} shows a modest edge on NIAH-MV at 16K (+1.0pp).

\begin{table}[t]
  \centering
  \caption{RULER results by task category at 4K/8K/16K context lengths.}
  \label{tab:ruler}
  \small
  \resizebox{0.8\textwidth}{!}{
  \begin{tabular}{l ccc ccc ccc}
    \toprule
    & \multicolumn{3}{c}{\textbf{NIAH-MV}} & \multicolumn{3}{c}{\textbf{CWE}} & \multicolumn{3}{c}{\textbf{VT}} \\
    \cmidrule(lr){2-4} \cmidrule(lr){5-7} \cmidrule(lr){8-10}
    \textbf{Method} & 4K & 8K & 16K & 4K & 8K & 16K & 4K & 8K & 16K \\
    \midrule
    Pretrained       & 98.4 & 95.1 & 94.0 & 98.0 & 83.2 & 66.7 & 99.6 & 96.5 & 93.7 \\
    Standard SFT     & 97.6 & 94.6 & 94.6 & 97.5 & 85.4 & 72.9 & 100.0 & 97.4 & 94.6 \\
    \textsc{FocuSFT} & \textbf{98.9} & \textbf{95.4} & \textbf{95.6} & \textbf{97.8} & \textbf{88.8} & \textbf{81.1} & 100.0 & \textbf{97.5} & 94.4 \\
    \bottomrule
  \end{tabular}
  }
\end{table}

\minisection{Downstream tasks}
\Cref{tab:downstream} reports results on LongBench QA and GPQA.
On LongBench, \textsc{FocuSFT} improves average F1 by +2.4pp, with the largest gain on MultifieldQA (+5.2pp), which requires cross-document evidence aggregation.
On GPQA, \textsc{FocuSFT} achieves 19.4\% vs.\ 15.6\% pass@1 (+3.8pp), suggesting that training-time attention improvements can transfer to complex agentic reasoning over long heterogeneous contexts.

\begin{table}[t]
  \caption{Downstream evaluation: LongBench QA (F1, 8K context) and GPQA Diamond (agentic tool use, $n{=}32$).}
  \label{tab:downstream}
  \centering
  \small
  \begin{tabular}{l ccccc cc}
    \toprule
    & \multicolumn{5}{c}{\textbf{LongBench QA}} & \multicolumn{2}{c}{\textbf{GPQA}} \\
    \cmidrule(lr){2-6} \cmidrule(lr){7-8}
    \textbf{Method} & HotpotQA & Multifield & Narrative & Qasper & Avg & pass@1 & pass@32 \\
    \midrule
    Standard SFT     & 10.7 & 28.6 & 8.7  & 12.2 & 15.0 & 15.6 & 78.8 \\
    \textsc{FocuSFT} & \textbf{11.6} & \textbf{33.8} & \textbf{11.3} & \textbf{12.8} & \textbf{17.4} & \textbf{19.4} & \textbf{80.8} \\
    \bottomrule
  \end{tabular}
\end{table}

\subsection{Ablation Study}
\label{sec:ablation}

\begin{table}[t]
  \centering
  \caption{Ablation study on BABILong. The 2$\times$2 design isolates bilevel optimization and bidirectional context attention.}
  \label{tab:ablation}
  \small
  \begin{tabular}{l cc cccc}
    \toprule
    \textbf{Method} & \textbf{Bilevel} & \textbf{Bidir} & \textbf{4K} & \textbf{8K} & \textbf{16K} & \textbf{32K} \\
    \midrule
    Standard SFT           & \texttimes & \texttimes & 70.2 & 62.6 & 47.4 & 33.2 \\
    SFT + Bidir.           & \texttimes & \checkmark & 65.4 & 57.6 & 46.0 & 33.0 \\
    Causal Bilevel         & \checkmark & \texttimes & 82.4 & 72.6 & 56.6 & 39.0 \\
    \textsc{FocuSFT}       & \checkmark & \checkmark & \textbf{84.4} & \textbf{72.8} & \textbf{57.6} & \textbf{42.8} \\
    \bottomrule
  \end{tabular}
\end{table}

\Cref{tab:ablation} presents the 2$\times$2 factorial results.
Bilevel optimization is the primary driver of improvement, accounting for +12.2, +10.0, +9.2, and +5.8pp over Standard SFT at 4K/8K/16K/32K, supporting the hypothesis that the inner-loop parametric memory concentrates attention on salient context during training.
Bidirectional context attention alone (SFT + Bidir.) actually degrades performance by 4--5pp at shorter lengths: without the inner loop to leverage the richer representations, the train--eval mismatch between bidirectional training and causal inference introduces a distribution shift.
However, when combined with bilevel optimization, bidirectional attention provides an additional +2.0, +0.2, +1.0, and +3.8pp over Causal Bilevel.
The gap widens at 32K, where attention dilution is most severe and bidirectional encoding enables fuller aggregation of dispersed evidence; the combined gain of +9.6pp exceeds the sum of individual effects (+5.8 and $-$0.2), indicating a positive interaction between the two components.

\begin{figure}[t]
  \centering
  % \vspace{-2cm}
  \begin{minipage}[t]{0.45\textwidth}
    \centering
    \includegraphics[width=\textwidth]{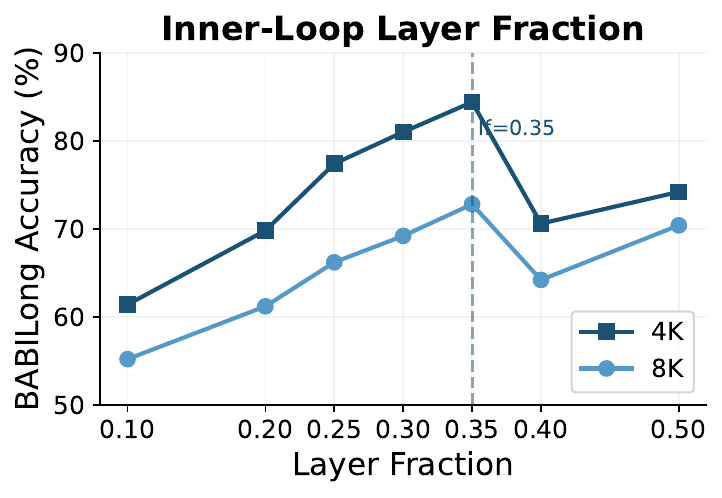}
    \caption{BABILong accuracy vs.\ inner-loop layer fraction. Performance peaks at lf=0.35, balancing memory capacity and base model stability.}
    \label{fig:layer_fraction}
  \end{minipage}
  \hfill
  \begin{minipage}[t]{0.45\textwidth}
    \centering
    \includegraphics[width=\textwidth]{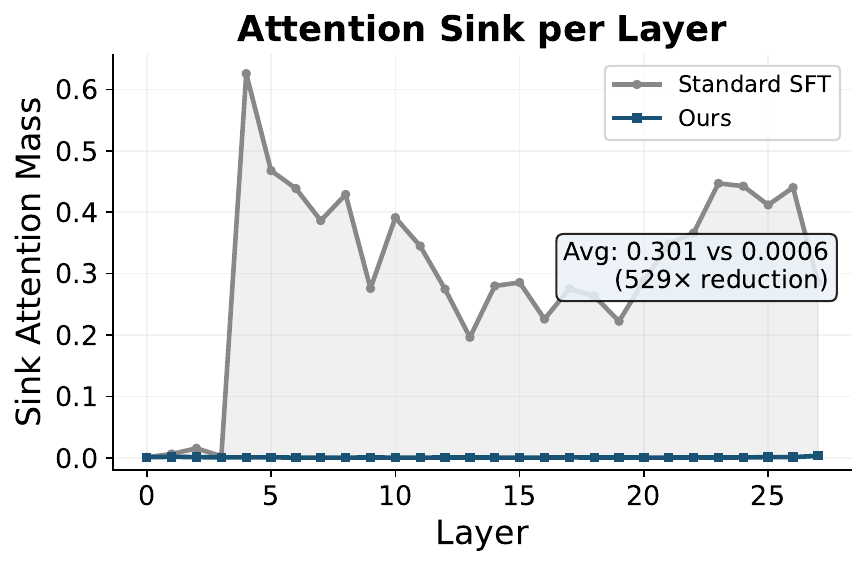}
    \caption{Attention sink mass per layer. Standard SFT exhibits a pervasive sink (avg 30.1\%); \textsc{FocuSFT} reduces it to 0.06\% (529$\times$ reduction).}
    \label{fig:sink_per_layer}
  \end{minipage}
\end{figure}

\minisection{Layer fraction sensitivity}
\Cref{fig:layer_fraction} shows BABILong accuracy as a function of the fraction of layers receiving LoRA adaptation in the inner loop.
Performance exhibits a clear inverted-U shape, peaking at $\text{lf}{=}0.35$.
Too few adapted layers (lf$\le$0.20) under-capacitate the parametric memory, limiting its ability to encode context-specific representations.
Too many (lf$\ge$0.40) cause the inner loop to over-specialize, disrupting the base model's pretrained representations and degrading the outer-loop gradient signal.

\minisection{Other bilevel hyperparameters}
The number of inner steps $K$ and the inner learning rate exhibit similar sensitivity.
$K{=}2$ is optimal; increasing to $K{=}3$ causes the inner loop to overfit to the current training sample, producing an outer gradient that no longer reflects general context utilization and leading to significant performance degradation.
The inner learning rate must be sufficiently high (we use 1.0) to enable meaningful adaptation within only two gradient steps; substantially lower values underadapt, while excessively high values destabilize training.
These interactions are consistent with the general principle that the inner loop should form a rough context sketch rather than memorize the sample.

\minisection{Training efficiency}
\Cref{tab:efficiency} reports training cost.
The bilevel inner loop is the dominant overhead (1.52$\times$); bidirectional attention adds negligible cost since it only changes the mask without additional parameters.
\textsc{FocuSFT} totals 1.71$\times$ wall time (a modest premium for +14.2pp on BABILong) and incurs zero inference overhead, as the inner-loop adapters are discarded after training.

\begin{table}[t]
  \centering
  \caption{Training efficiency (8$\times$ GPU, 469 steps). All methods share the same inference cost.}
  \label{tab:efficiency}
  \small
  \begin{tabular}{l ccc c}
    \toprule
    \textbf{Method} & \textbf{s/step} & \textbf{Wall Time} & \textbf{Overhead} & \textbf{BABILong 4K} \\
    \midrule
    Standard SFT     & 3.64 & 0.47h & 1.00$\times$ & 70.2 \\
    SFT + Bidir.     & 3.64 & 0.47h & 1.00$\times$ & 65.4 \\
    Causal Bilevel   & 5.52 & 0.72h & 1.52$\times$ & 82.4 \\
    \textsc{FocuSFT} & 6.21 & 0.81h & 1.71$\times$ & \textbf{84.4} \\
    \bottomrule
  \end{tabular}
\end{table}

\subsection{Attention Analysis}
\label{sec:attn_analysis}

\Cref{sec:training_dilution} identified training-time attention dilution in standard SFT.
Here we examine how \textsc{FocuSFT} reshapes these patterns.

\minisection{Per-layer sink reduction}
\Cref{fig:sink_per_layer} shows that the attention sink under standard SFT is not confined to a few layers but is pervasive across all 28 layers.
\textsc{FocuSFT} reduces the per-layer sink mass to 0.06\% (a 529$\times$ reduction), confirming that bidirectional context attention removes the causal asymmetry that sustains the sink mechanism throughout the network.

\minisection{Attention budget redistribution}
With the sink eliminated, the freed attention budget is redirected to semantically relevant content (\Cref{fig:budget_detail}).
System and user prompt attention rises from 7.0\% to 27.0\%, and tool response attention from 6.5\% to 14.3\%, yielding 3.1$\times$ higher total context engagement (13.5\% $\to$ 41.4\%).
The positional profile (\Cref{fig:positional_attn}) reflects this shift: the spike at position 0 disappears and is replaced by content-driven peaks aligned with semantically meaningful turns.

\minisection{Structural transformation}
The attention heatmaps (\Cref{fig:attn_heatmap}) reveal the structural consequence.
Under standard SFT, a uniform sink column dominates across all query positions, leaving the middle context substantially underattended.
Under \textsc{FocuSFT}, attention is visibly spread across the middle context positions that were previously ignored: response queries now attend to content tokens throughout the sequence, not just the initial positions.
The heatmap also reveals the multi-turn dialogue structure: context turns exhibit bidirectional attention blocks, while assistant responses maintain causal masking.
This shift from positionally concentrated to content-distributed attention is the mechanism through which the budget redistribution above translates into improved long-context utilization.

\section{Related Work}
\label{sec:related}

\minisection{Long-context architectures}
Extending the context window of transformer-based LLMs has been an active area of research.
Positional encoding improvements such as RoPE scaling~\cite{liu2023scaling, ding2024longrope, peng2023yarn} and ALiBi~\cite{press2021train} extrapolate to longer sequences, while distributed mechanisms like ring attention~\cite{liu2023ring} address memory constraints.
Sparse attention patterns~\cite{child2019generating, beltagy2020longformer, yuan2025native} reduce the quadratic cost while preserving long-range connectivity, and Differential Transformer~\cite{ye2024differential} subtracts two softmax maps to cancel attention noise.
These methods expand context \emph{visibility} or efficiency but do not address the attention dilution that limits context \emph{utilization}~\cite{hsieh2024ruler, liu2024lost}.
\textsc{FocuSFT} is orthogonal to these architectural advances: it keeps the standard attention mechanism intact and modifies only the training procedure.

\minisection{Inference-time long-context methods}
A growing family of methods address long-context failures at inference time.
``Found in the Middle''~\cite{hsieh2024found} applies post-hoc positional calibration; DySCO~\cite{ye2026dysco} dynamically up-weights retrieval-head scores during decoding.
Test-time training (TTT) approaches~\cite{pmlr-v119-sun20b, bansal2025let, tandon2025end} are methodologically closest to \textsc{FocuSFT}: they also adapt model parameters on the input context via gradient steps before generation.
The key difference is that TTT operates at inference time on each new input, incurring per-sample compute overhead, while \textsc{FocuSFT} uses the same inner-loop mechanism during training only, producing a base model that requires no additional inference-time computation.

\minisection{Long-context fine-tuning}
LongLoRA~\cite{longlora} enables efficient long-context fine-tuning via shifted sparse attention during training with full attention at inference, and LongAlpaca~\cite{long-alpaca} provides instruction-following data for long-context adaptation.
LongAlign~\cite{bai2024longalign} further studies long-context alignment data construction and training efficiency.
These approaches increase the model's ability to process longer inputs but do not specifically address how attention budget is distributed during fine-tuning.
\textsc{FocuSFT} targets a complementary axis: rather than extending the context window, it improves how the model uses the context already within its window during SFT.

\minisection{Bilevel optimization and fast weights}
Bilevel optimization is well established in meta-learning~\cite{thrun1998learning, finn2017model, nichol2018first, rajeswaran2019meta}, where MAML~\cite{finn2017model} uses inner-loop adaptation to learn task-generalizing initializations.
Recent work has applied bilevel optimization to LLM data reweighting~\cite{pan2025scalebio}.
Fast weights~\cite{hinton1987using, ba2016using} maintain a secondary set of parameters updated on a faster timescale; Fast Weight Layers~\cite{clark2022meta} express this as linear attention, and MemoryLLM~\cite{memoryllm} integrates context into model parameters for knowledge retention.
\textsc{FocuSFT} aims at a different goal: not cross-task generalization or knowledge storage, but \emph{intra-sequence} attention sharpening during long-context SFT.

\section{Limitations}
\label{sec:limitations}

\textsc{FocuSFT} introduces a 1.71$\times$ training time overhead due to the inner-loop gradient steps (\Cref{tab:efficiency}), though inference cost is unchanged.
Our experiments use a single model family (Qwen2.5-7B) and a fixed training corpus of 3K samples; scaling behavior across model sizes and larger data regimes remains to be explored.
The bilevel formulation currently targets the SFT stage; integration with reinforcement learning from human feedback (RLHF) or direct preference optimization (DPO) is a natural extension but has not been investigated.
% Finally, while we demonstrate gains on multi-turn reasoning and QA benchmarks, evaluation on additional long-context application domains (e.g., repository-level code understanding, long-document summarization) would further characterize the method's generality.

\section{Conclusion}
\label{sec:conclusion}

We have argued that the widely observed gap between long-context \emph{visibility} and \emph{utilization} in LLMs is not solely an inference-time problem; it is rooted in training, where positional biases and attention sinks starve content tokens of attention, corrupting the learning signal itself.
To address this training-time attention dilution, we introduced \textsc{FocuSFT}, a bilevel optimization framework.
Through fast-weight adaptation in the inner loop, \textsc{FocuSFT} forms a parametric memory that concentrates attention on semantically relevant content; the outer loop then performs SFT conditioned on this sharpened representation.
Bidirectional context attention reduces the causal asymmetry that gives rise to attention sinks, while inner-outer consistency aligns the sharpened representations with downstream use.
Together, these mechanisms help mitigate the vicious cycle in which diluted training produces models with poor long-context capabilities.
% Empirically, \textsc{FocuSFT} improves BABILong reasoning accuracy by up to +14.2pp, boosts CWE aggregation by +8.2pp at 16K context, and raises GPQA agentic pass@1 by +3.8pp, all with zero inference overhead.

\newpage
{
\small
\bibliographystyle{neurips}
\bibliography{ref/Top,ref/main}
}

\newpage
\appendix
\section{Additional Experimental Details}
\label{app:details}

\paragraph{Training hyperparameters.}
\Cref{tab:hyperparams} lists all hyperparameters for the main \textsc{FocuSFT} configuration.

\begin{table}[h]
  \centering
  \caption{Hyperparameters for \textsc{FocuSFT} training.}
  \label{tab:hyperparams}
  \small
  \begin{tabular}{ll}
    \toprule
    \textbf{Parameter} & \textbf{Value} \\
    \midrule
    \multicolumn{2}{l}{\textit{Outer loop (SFT)}} \\
    Base model & Qwen2.5-7B \\
    Optimizer & AdamW ($\beta_1{=}0.9$, $\beta_2{=}0.999$) \\
    Learning rate & $1 \times 10^{-5}$ \\
    LR schedule & Cosine with 10\% linear warmup \\
    Weight decay & 0.01 \\
    Max gradient norm & 1.0 \\
    Epochs & 5 \\
    Effective batch size & 32 (8 GPUs $\times$ 1 $\times$ GA 4) \\
    Max sequence length & 4096 \\
    Precision & BF16 \\
    \midrule
    \multicolumn{2}{l}{\textit{Inner loop (parametric memory)}} \\
    Inner steps ($K$) & 2 \\
    Inner learning rate & 1.0 \\
    Inner gradient clip & 1.0 \\
    Adapter type & LoRA \\
    LoRA rank & 32 \\
    LoRA $\alpha$ & 64 \\
    LoRA dropout & 0.0 \\
    Target modules & \texttt{gate\_proj}, \texttt{up\_proj}, \texttt{down\_proj} \\
    Layer fraction & 0.35 (top 10 of 28 layers) \\
    Attention mode & Bidirectional context (GLM-style) \\
    \midrule
    \multicolumn{2}{l}{\textit{Other}} \\
    Seed & 1234 \\
    Gradient checkpointing & Enabled \\
    Training samples & 3,000 \\
    Hardware & 8$\times$ NVIDIA GPU \\
    \bottomrule
  \end{tabular}
\end{table}

\paragraph{GPQA evaluation.}
We evaluate GPQA Diamond using the Open-AgentRL~\cite{yu2025demystify} framework with multi-turn tool-use rollouts in Hermes chat format.
Each of the 198 problems receives $n{=}32$ independent rollouts with temperature 1.0 and top-$p$ 0.6, allowing up to 16 assistant turns per episode.
We report pass@1 (majority vote) and pass@32 (oracle across all rollouts).

\section{Additional Results}
\label{app:results}

\subsection{Per-Task BABILong Breakdown}
\label{app:babilong_pertask}

\Cref{tab:babilong_pertask} shows per-task BABILong accuracy at 4K and 16K context lengths.
The largest gains appear on QA2 (two-fact reasoning, +26pp at 4K) and QA3 (temporal reasoning, +31pp at 4K, +30pp at 16K), both of which require the model to locate and connect multiple dispersed facts across the context, precisely the scenario where attention dilution is most damaging.

\begin{table}[h]
  \centering
  \caption{Per-task BABILong accuracy. QA2 and QA3 require multi-hop reasoning over dispersed facts.}
  \label{tab:babilong_pertask}
  \small
  \begin{tabular}{ll cccccc}
    \toprule
    & & \multicolumn{3}{c}{\textbf{4K}} & \multicolumn{3}{c}{\textbf{16K}} \\
    \cmidrule(lr){3-5} \cmidrule(lr){6-8}
    \textbf{Task} & \textbf{Type} & SFT & FocuSFT & $\Delta$ & SFT & FocuSFT & $\Delta$ \\
    \midrule
    QA1 & Single fact  & 78.0 & 84.0 & +6.0  & 57.0 & 52.0 & $-$5.0 \\
    QA2 & Two facts    & 51.0 & 77.0 & \textbf{+26.0} & 19.0 & 31.0 & +12.0 \\
    QA3 & Temporal     & 53.0 & 84.0 & \textbf{+31.0} & 37.0 & 67.0 & \textbf{+30.0} \\
    QA4 & Spatial      & 82.0 & 89.0 & +7.0  & 63.0 & 69.0 & +6.0 \\
    QA5 & Argument     & 87.0 & 88.0 & +1.0  & 61.0 & 69.0 & +8.0 \\
    \midrule
    \multicolumn{2}{l}{Average} & 70.2 & \textbf{84.4} & +14.2 & 47.4 & \textbf{57.6} & +10.2 \\
    \bottomrule
  \end{tabular}
\end{table}

\subsection{RULER 2$\times$2 Ablation}
\label{app:ruler_ablation}

\Cref{tab:ruler_ablation} extends the BABILong ablation (\Cref{tab:ablation}) to RULER, confirming the same pattern: bilevel optimization provides the primary benefit on aggregation tasks (CWE), while bidirectional attention alone slightly degrades performance at 16K.

\begin{table}[h]
  \centering
  \caption{RULER 2$\times$2 ablation (average accuracy across NIAH-MV, CWE, VT).}
  \label{tab:ruler_ablation}
  \small
  \begin{tabular}{l cc ccc}
    \toprule
    \textbf{Method} & \textbf{Bilevel} & \textbf{Bidir} & \textbf{Avg@4K} & \textbf{Avg@8K} & \textbf{Avg@16K} \\
    \midrule
    Standard SFT     & \texttimes & \texttimes & 98.3 & 92.5 & 87.3 \\
    SFT + Bidir.     & \texttimes & \checkmark & 98.5 & 92.1 & 84.9 \\
    Causal Bilevel   & \checkmark & \texttimes & 97.9 & 91.9 & 90.3 \\
    \textsc{FocuSFT} & \checkmark & \checkmark & \textbf{98.9} & \textbf{93.9} & \textbf{90.4} \\
    \bottomrule
  \end{tabular}
\end{table}

\subsection{Inference-Time Adaptation}
\label{app:inference_adapt}

A natural question is whether the inner-loop adapters can also be applied at inference time for additional gains.
We tested this by performing a single inner-loop gradient step on the test input before generation.
\Cref{tab:inference_adapt} shows that inference-time adaptation does not consistently improve over the base \textsc{FocuSFT} model and slightly hurts on most benchmarks.

\begin{table}[h]
  \centering
  \caption{Effect of inference-time adaptation. The bilevel-trained model already internalizes the attention-sharpening benefit; additional test-time adaptation is unnecessary.}
  \label{tab:inference_adapt}
  \small
  \begin{tabular}{l cccc}
    \toprule
    & \multicolumn{4}{c}{\textbf{BABILong}} \\
    \cmidrule(lr){2-5}
    \textbf{Method} & 4K & 8K & 16K & 32K \\
    \midrule
    \textsc{FocuSFT}              & \textbf{84.4} & \textbf{72.8} & \textbf{57.6} & \textbf{42.8} \\
    \textsc{FocuSFT} + Inf.\ Adapt. & 83.4 & 71.6 & 53.4 & 41.6 \\
    \bottomrule
  \end{tabular}
\end{table}

\textsc{FocuSFT} therefore incurs zero inference overhead: the bilevel training procedure improves the base model weights directly, and the inner-loop adapters are discarded after training.

\section{Attention Visualization Details}
\label{app:attn_vis}

The attention heatmaps (\Cref{fig:attn_heatmap}) and per-layer analysis (\Cref{fig:sink_per_layer}) are computed from a representative 4096-token multi-turn agentic sample containing 5 context turns (system prompt + 4 tool responses) and 5 assistant response turns.
Attention weights are extracted from layer 14 (middle of the network) for heatmaps, and averaged across all attention heads.
The sink mass metric sums attention weight on positions [0:5] from all response-query positions.

\end{document}